\documentclass[letterpaper, 10 pt, journal, twoside]{ieeetran}

\IEEEoverridecommandlockouts                              

\usepackage[utf8]{inputenc} 
\usepackage[T1]{fontenc}    
\usepackage{booktabs}       
\usepackage{amsfonts}       
\usepackage{nicefrac}       
\usepackage{microtype}      
\usepackage{graphicx}
\usepackage{amsmath}
\usepackage{color}
\usepackage{array}

\usepackage{algorithm}
\usepackage{algpseudocode}

\usepackage{soul}  

\newcommand{\mbs}[1]{\ensuremath{\boldsymbol{#1}}}

\newcommand{\cX}{\mathcal{X}}

\newcommand{\ex}{\mathbb{E}}

\newcommand{\RR}{\mathbb{R}}

\newcommand{\thetav}{\mbs{\theta}}

\newcommand{\av}{\mathbf{a}}

\newcommand{\kv}{\mathbf{k}}

\newcommand{\s}{\mathbf{s}}

\newcommand{\x}{\mathbf{x}}
\newcommand{\y}{\mathbf{y}}

\newcommand{\I}{\mathbf{I}}
\newcommand{\K}{\mathbf{K}}

\newcommand{\X}{\mathbf{X}}

\newcommand{\N}{\mathcal{N}}

\title{\LARGE \bf
Robust policy search for robot navigation
}

\author{Javier Garcia-Barcos and Ruben Martinez-Cantin
\thanks{Manuscript received: October, 16, 2020; Revised January, 14, 2021; Accepted February, 3, 2021.}
\thanks{This paper was recommended for publication by
Editor Dana Kulic upon evaluation of the Associate Editor and Reviewers’
comments.
This work was partly supported by projects RTI2018-096903-B-I00 (MINECO/FEDER) and RED2018-102511-T.} 
\thanks{Javier Garcia-Barcos and Ruben Martinez-Cantin are with the Instituto de Investigacion en Ingenieria de Aragon (I3A), University of Zaragoza, Spain. {\tt\footnotesize {jgbarcos,rmcantin}@unizar.es}}%
\thanks{Digital Object Identifier (DOI): see top of this page.}
}

\begin{document}

\maketitle

\begin{abstract}

Complex robot navigation and control problems can be framed as policy search problems. However, interactive learning in uncertain environments can be expensive, requiring the use of data-efficient methods. Bayesian optimization is an efficient nonlinear optimization method where queries are carefully selected to gather information about the optimum location. This is achieved by a surrogate model, which encodes past information, and the acquisition function for query selection. Bayesian optimization can be very sensitive to uncertainty in the input data or prior assumptions. In this work, we incorporate both robust optimization and statistical robustness, showing that both types of robustness are synergistic. For robust optimization we use an improved version of \emph{unscented Bayesian optimization} which provides safe and repeatable policies in the presence of policy uncertainty. We also provide new theoretical insights. For statistical robustness, we use an adaptive surrogate model and we introduce the \emph{Boltzmann selection} as a stochastic acquisition method to have convergence guarantees and improved performance even with surrogate modeling errors. We present results in several optimization benchmarks and robot tasks.

\end{abstract}

\begin{IEEEkeywords}
Reinforcement Learning, Machine Learning for Robot Control, Optimization and Optimal Control, Probability and Statistical Methods
\end{IEEEkeywords}

\section{INTRODUCTION}

\IEEEPARstart{R}{obot} navigation in uncertain environments can be framed as a policy search problem \cite{williams1992simple}, a technique that has led to important achievements in robotics \cite{deisenroth2013survey,Peters06,Kohl04,levine2014learning}. Those results had been obtained using different flavours of gradient-based policy search, which might require a large number of trials and a good initialization to avoid suboptimal results in local minima. 
Trials for robotic applications come at a substantial cost, as each one requires to move and interact with a robot. Alternatively, high-performance robotic simulators still require a fair amount of computational resources. It is thus evident that sample efficiency is of paramount importance.

Active policy search uses Bayesian optimization to drive the search for optimality in an effective way. Bayesian optimization is a sample efficient method for nonlinear optimization that does not require gradients or good initialization to obtain global convergence \cite{shahriari2016taking}. The probabilistic nature of Bayesian optimization allows the use of partial or incomplete information, analogous to the stochastic gradient descent commonly used in classical policy search \cite{williams1992simple}. In fact, Bayesian optimization has been already used in some robotics and reinforcement learning setups, such as robot walking \cite{Lizotte2007,Calandra2015a,rai2019using}, control \cite{Tesch_2011_7370,Kuindersma01062013}, planning \cite{MartinezCantin09AR,marchant2014bayesian}, grasping \cite{nogueira2016unscented, castanheira2018finding,ActiveRewardLearning} and damage recovery \cite{Cully2015}.

\begin{figure}
  \centering
  \includegraphics[width=0.8\linewidth]{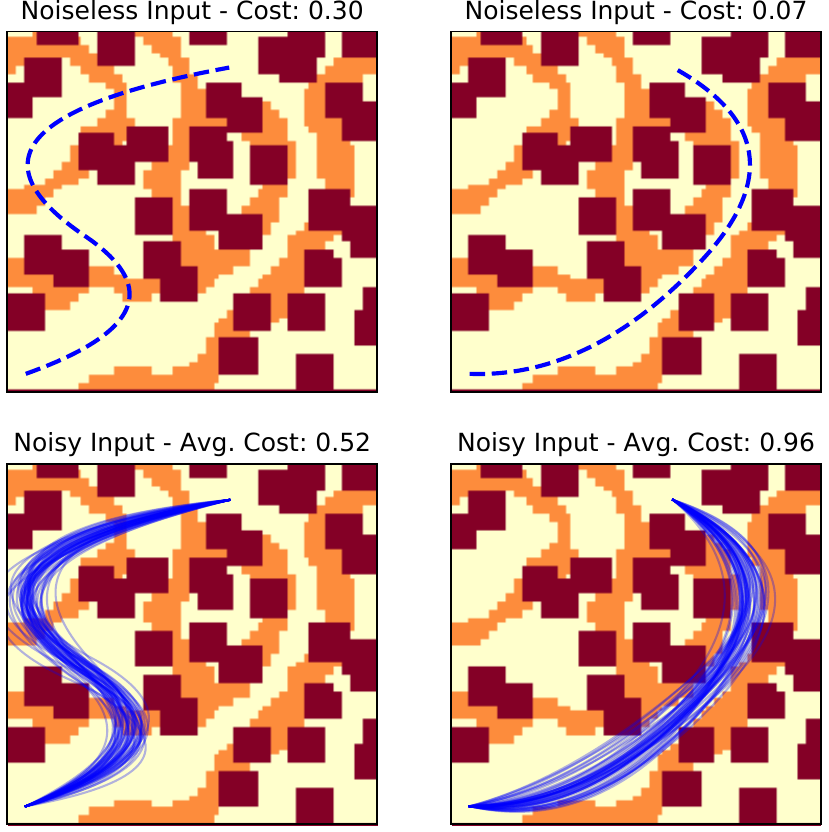}
  \caption{Path planning on uneven terrain with obstacles, with different trajectories displayed (left and right). The orange regions represent slopes with a higher traversing cost. The red rectangles are obstacles. Top: the desired trajectories (blue dashed line). Bottom: possible deviations (blue lines) from desired trajectories due to input noise. The right trajectory is more efficient without input noise. Once we take into account input noise, it becomes unsafe as it can collide with obstacles easily. The left trajectory is safer in the presence of input noise despite being less efficient.}
  \label{fig:rover_example}
\end{figure}

Bayesian optimization relies on a probabilistic surrogate model of the target function, typically a Gaussian process. In the original formulation, this model is incorporated for sample efficiency as it provides a \emph{memory} of previous trials \cite{Jones:1998}. The surrogate model can also be exploited for other purposes that can be useful in robotics or reinforcement learning scenarios, such as, guaranteeing a minimum outcome \cite{sui2015safe}, detect and remove outliers \cite{martinez2018practical} or incorporate prior information \cite{Cully2015}.


In this work, we are exploiting the probabilistic features of Bayesian optimization to provide \emph{robustness} to policy search. First, we imply robustness in the sense of robust optimization or robust control. The resulting policy should perform well even if the robot or agent is not able to follow the policy with enough precision.
For example, consider the navigation problem from Figure \ref{fig:rover_example}, although the right trajectory is shorter and cost-efficient, it is also riskier. If the trajectory uncertainty increases, as in the bottom plots, it becomes unsafe and incurs a higher cost on average. On the contrary, the left trajectory is longer and less efficient, but it is also safer. For a particular level of uncertainty, it becomes a safer and more efficient route on average. If we think on the cost function in terms of the policy parameters, the left trajectory lies in a smooth flat region while the right trajectory lies in a high variability region with a narrow valley. Intuitively, in the presence of location uncertainty or safety concerns, we want to avoid narrow optima where perturbations might push the solution to poor performance results. Instead, we focus on a broad optima where the solution is optimal even after perturbation. However, depending on the task and environmental conditions, the algorithm should be able to model and select between narrow and flat optimum regions. 
In \emph{robust optimization} and \emph{robust control}, the performance is usually maintained in a bounded region or set. In this work, we replace that region by a probability distribution (e.g.: Gaussian), because that is the typical representation for uncertainty in robot location. Thus, we focus on optimizing the averaged performance, while classical robust optimization optimizes the worst case scenario. We have developed a variant of Bayesian optimization that relies on the unscented transformation to consider the expected policy performance under uncertainty. One advantage of our Unscented Bayesian Optimization is that it can be used just for safety reasons (e.g.: if we are able to train in a simulator with perfect repeatability but we want to consider possible perturbations when executed in the real robot) or if the policy uncertainty comes from noisy or perturbed trials (e.g.: the policy parameters are perturbed during training). In the experiments in this paper, we consider the most challenging scenario of having perturbations also during training. 

\begin{figure}
  \smallskip \smallskip
  \centering
  \includegraphics[width=0.99\linewidth]{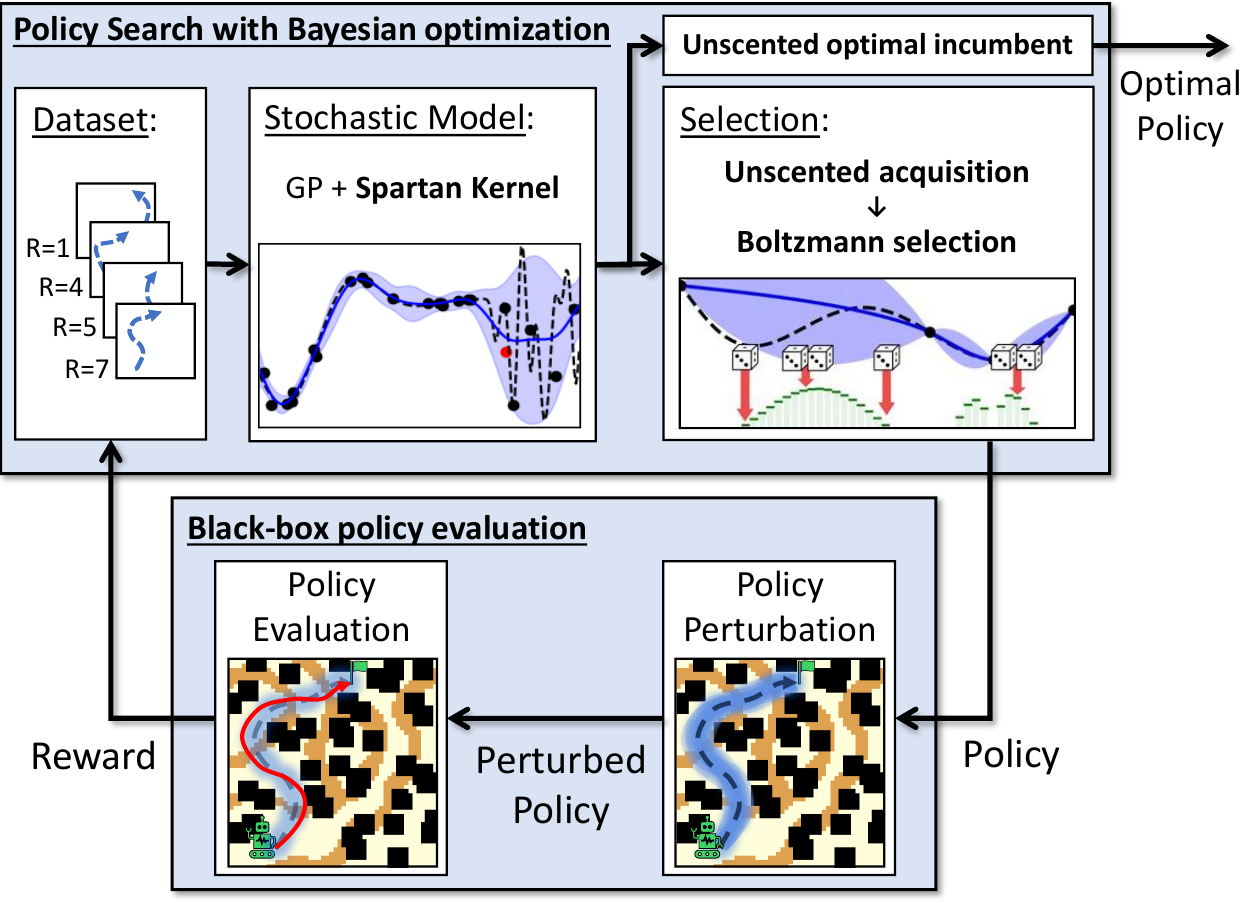}
  \caption{Diagram showing the different components of our approach, based on policy search with Bayesian optimization. It depicts the Bayesian optimization loop (top) applied to a policy search problem (bottom). The goal is to identify the most efficient policy by sequentially querying different policies and obtaining the corresponding reward. However, as the problem has policy uncertainty, a perturbed policy will be evaluated instead. We highlight (bold text) the elements that differ from a standard policy search with Bayesian optimization: the \textit{Spartan kernel} to model nonstationarity, the unscented transform applied in \textit{unscented optimal incumbent} and \textit{unscented acquisition}) to propagate the policy uncertainty and, instead of greedy acquisition function maximization, \textit{Boltzmann selection} sampling to improve exploration in the presence of surrogate modeling errors. 
  }
  \label{fig:approach_diagram}
\end{figure}

It has been studied that reward functions found in robotics might be difficult to approximate by typical surrogate models in Bayesian optimization, resulting in unreliable performance of the optimizer \cite{MartinezCantin17icra}. Furthermore, policy perturbations during training, as discussed before, can also be problematic for the surrogate model and the optimizer performance. Therefore, in this work, we have also included another layer of \emph{robustness} to our proposal.
\emph{Robust statistics} deals with statistical methods that perform reasonably well even when the underlying assumptions are somehow violated. For example, convergence of Bayesian optimization methods is based on the assumption that the target function \emph{belongs} to the reproducing kernel Hilbert space spanned by the Gaussian process kernel. Bayesian optimization relies on optimal decision theory to actively select the next informative trial. When combined with an inadequate model, this can lead to poor results and lack of convergence. 
In this work, we employ several strategies to provide robustness in a statistical sense. First, we use an adaptive kernel for nonstationary environments \cite{MartinezCantin17icra}. This provides a much more flexible surrogate model to accommodate a larger set of target functions, such as common reward functions. Our method also selects new trials based on \emph{Boltzmann selection} that can be robust to surrogate modeling errors \cite{ijcai2019-0327}. The intuition behind the Boltzmann selection is to select new policies based on a softmax of the acquisition function, instead of the standard optimum. An advantage of this approach is that it can be used to derive convergence bounds without assuming artificially injected exploration. Another advantage of the Boltzmann selection is that they trivially allow to perform distributed Bayesian optimization in a multi-robot setup or using a simulator.
Figure \ref{fig:approach_diagram} shows the different components of our approach. 

We present a new architecture for robust efficient policy search and we provide new theoretical analysis and insights of the methods employed. Specifically, we introduce the first robust policy search both in terms of robust optimization and statistical robustness. Furthermore, policy search is performed using episodic Bayesian optimization, which requires a small number of trials to obtain optimal results. Our extensions maintain the data efficiency of standard Bayesian optimization. Our algorithm combines two novel methods: \textbf{Unscented Bayesian Optimization} and \textbf{Boltzmann selection} (preliminary versions were published in \cite{nogueira2016unscented, ijcai2019-0327}) with other ingredients: expected improvement and Gaussian processes with adaptive kernels to guarantee optimal and stable solutions even under broken assumptions, biased priors, query perturbations, etc. As secondary contributions: 1) we design the first fully distributed robust policy search algorithm, which allows parallel evaluations in multirobot systems and simulators. 2) we provide a theoretical interpretation of the unscented Bayesian optimization as an integrated response method with polynomial complexity and a formulation based on the scaled unscented transform, which provides more flexibility to define the safety/stability region.



\section{BACKGROUND}

\subsection{Active policy search\label{sec:policysearch}}
Policy search consists of finding the optimal parameters $\x^*$ of a policy $\pi_\x(\av_k|\s_k)$ with respect to the expected return $U^\pi$, denoted $\ex[U^{\pi}] = \ex [\sum_{k=1}^M \gamma^{k} R^\pi(\s_k, \av_k)]$. Here, $a$ denotes the action, $s$ the state, $\gamma$ the discounted factor and $M$ the episode length. Without loss of generality, we assume finite horizon on episodic policy search. The expectation is under the policy and the system dynamics which together form a distribution over trajectories $\tau$. If we use an episodic formulation, such as REINFORCE \cite{williams1992simple}, the expectation is usually approximated from Monte-Carlo rollouts $\tau^{(i)} \sim \tau$ of the robot trajectory. In this setup, finding the optimal policy parameters can be framed as a pure optimization problem, where the objective function is then computed as:
\begin{equation}
  \label{eq:reward}
  \begin{split}
  f(\x) &= \ex_\tau[U^{\pi}] \approx \sum_{i=1}^N \sum_{k=1}^M \gamma^{k} R\left(\tau_k^{(i)}\right)  \\
  \x^* &= \arg \max_{\x \in \mathcal{X}} \ex_\tau[U^{\pi}]
  \end{split}
  \end{equation}
  
where $\x^*$ are the parameters of the optimal policy $\pi^* = \pi_{\x^*}$, $R(\tau_k^{(i)})$ is the instantaneous reward at time step $k$ following rollout $\tau^{(i)}$ and $N$ is the number of rollouts. Active policy search \cite{MartinezCantin07RSS} computes the optimal policy parameters using Bayesian optimization. Similarly to stochastic gradient descent in gradient-based policy search, Bayesian optimization can directly be applied to stochastic optimization thanks to the probabilistic surrogate model \cite{Huang06}. Therefore, the expectation in equation \eqref{eq:reward} can be approximated with a small batch of rollouts or even a single episode. Algorithm \ref{al:apl} summarized the active policy search strategy. Section \ref{sec:bayesianoptimization} details the steps of updating the surrogate model and generating the next set of policy parameters $\x_{t+1}$.

\begin{algorithm}
\renewcommand{\algorithmicrequire}{\textbf{Input:}}
\caption{Active Policy Search}\label{al:apl}
\begin{algorithmic}[1]
\Require Optimization budget $T$
  \State Initialize $\x_1$ based on a low discrepancy sequence.
  \For {each optimization iteration $t$ until budget $T$}:
   \State Generate episode $\tau_{\x_t} \sim \{s_0,a_0,R_1,s_1,a_1,\ldots\}$
   \State $y_t \leftarrow \sum_{k=1}^M \gamma^{k} R_k$
   \State Add $(\x_t, y_t)$ to surrogate model with equation \eqref{eq:predgp}
   \State Generate $\x_{t+1}$ using equation \eqref{eq:acquisition}
\EndFor

\end{algorithmic}
\end{algorithm}

\subsection{Bayesian optimization} \label{sec:bayesianoptimization}

Bayesian optimization is a framework that aims to efficiently optimize noisy, expensive, blackbox functions. It uses two distinct components: a \emph{probabilistic surrogate model} $p(f)$ that learns the properties and features of the target function using previously evaluated observations and an \emph{acquisition function} $\alpha(x,p(f))$ that, based on the surrogate model, builds an utility function which rates how promising a subsequent query could be. Although our contributions are agnostic to these choices, for the remainder of the paper, the discussion and results are based on the use of a \emph{Gaussian process} as the surrogate model and the \emph{expected improvement} as the acquisition function because they are the most commonly used in the literature due to their excellent performance in a large variety of problems.

Formally, Bayesian optimization attempts to find the global optima of an expensive unknown function $f:\cX\to\RR$ over some domain $\cX\subset\RR^d$ by sequentially performing queries. At iteration $t$, all previously observed values $\y=y_{1:t}$ at queried points $\X=\x_{1:t}$ are used to learn a probabilistic surrogate model $p(f|y_{1:t},\x_{1:t})$. Typically, the next query $\x_{t+1}$ is then determined by greedily optimizing the acquisition function in $\cX$:
\begin{equation}
  \label{eq:acquisition}
 \x_{t+1} = \arg \max_{\x \in \cX} \alpha\left(\x, p(f\;|\;y_{1:t},\x_{1:t})\right)
\end{equation}
although we will replace the greedy selection in Section \ref{sec:offpol}.

\paragraph{Surrogate Model\label{sec:gp}} The most common surrogate model is the Gaussian process (GP). For the remainder of the paper we consider a GP with zero mean and kernel $k:\cX\times\cX\to\RR$. 
The GP posterior model allows predictions at query points $\x_q$ which are normally distributed $y_q \sim \mathcal{N}(\mu(\x_q), \sigma^2(\x_q))$, such that:
\begin{equation} 
    \label{eq:predgp}
    \begin{split}
    \mu(\x_q) & = \kv(\x_q)^T\K^{-1}\y \\
    \sigma ^2 (\x_q) & = k(\x_q, \x_q) - \kv(\x_q)^T \K^{-1} \kv(\x_q)
    \end{split}
\end{equation}
where $\kv(\x_q) = \left[k(\x_q,\x_i)\right]_{\x_i \in \X}$ and $\K = \left[\kv(\x_i,\x_j)\right]_{\x_i,\x_j \in \X} + \mathbf{I}\sigma^2_n$. For the kernel, we have used the Spartan kernel which provides robustness to nonstationary function and improves convergence, which has been shown to be critical for reinforcement learning problems \cite{martinez2018funneled}. 

\paragraph{Kernel function\label{sec:kernel}} The Spartan kernel \cite{martinez2018funneled} is the combination of several local kernels $k_l$ applied over moving regions \emph{defined by weighting functions } $\omega_l(\x'| \thetav_{p})$, with a global kernel $k_g$ for the rest of the space \emph{defined by weight} $\omega_{g}(\x)$. Each weighting functions follow a normal distribution:
\begin{equation}
\begin{split}
  \label{eq:weights}
    \omega_{g} &= \N\left(\psi, \I \sigma^2_{g} \right), \\
    \omega_{l} &=  \N\left(\thetav_{p}, \I \sigma^2_{l} \right) \qquad \forall \; l = 1\ldots M
 \end{split}
\end{equation}
where $\psi$ and $\thetav_{p}$ can be seen as the center of the influence region of each kernel while $\sigma_{g}$ and $\sigma_{l}$ can be interpreted as the size of each area of influence. The regions of the local kernels are centered in a single point $\thetav_{p}$ with multiple diameters, creating a funnel structure. In order to achieve smooth interpolation between regions, we use normalized weights $\lambda_j(\x) = \sqrt{\omega_j(\x)/\sum_p \omega_p(\x)}$:
\begin{equation}
\begin{split}
  \label{eq:spartan}
  k_{S}(\x, \x'| \thetav_{S}) &= \lambda_{g}(\x) \lambda_{g}(\x') k_{g}(\x,\x'| \thetav_g) \\& + \sum_{l=1}^M \lambda_l(\x| \thetav_{p}) \lambda_l(\x'| \thetav_{p}) k_l(\x,\x'| \thetav_l)
\end{split}
\end{equation}
In the experiments, we have used Mat{\'e}rn kernels with automatic relevance determination for $k_g$ and $k_l$ \cite{rasmussen2006}. For the local kernels, we estimate the center of the funnel structure $\thetav_p$ based on the data gathered. Thus, we consider $\thetav_p$ as part of the hyperparameters jointly with the Mat{\'e}rn hyperparmeters $\thetav_{g}$ and $\thetav_{l}$:
\begin{eqnarray}
  \label{eq:spakernel}
\thetav_{S}=[\thetav_{g}, \thetav_{l_1},\ldots,\thetav_{l_M}, \thetav_{p}]
\end{eqnarray}
 
\paragraph{Hyperparameter estimation\label{sec:mcmc}} In many GP applications, including Bayesian optimization, kernel hyperparameters are estimated using the \emph{empirical Bayes} approach. In that case, a point estimate like the maximum likelihood or maximum a posteriori is used, resulting in an overconfident estimate of the GP uncertainty \cite{rasmussen2006}. Instead, we use a \textit{fully Bayesian} approach based on Markov chain Monte Carlo (MCMC) to generate a set of samples $\{\thetav_i\}_{i=1}^N$ with $\thetav_i \sim p(\thetav | \y, \X)$. In particular, we use the slice sampling algorithm which has already been used successfully in Bayesian optimization \cite{Snoek2012}. 

\paragraph{Acquisition Function\label{sec:ei}} The expected improvement (EI) \cite{Mockus78} is a standard acquisition function defined in terms of the query improvement at iteration $t$ and is defined as:
\begin{equation}
	\label{eq:ei}
	EI_t(\x) = \left(\rho_t - \mu_t\right) \Phi(z_t) + \sigma_t \phi(z_t)
\end{equation}

where $\phi$ and $\Phi$ are the corresponding Gaussian probability density function (PDF) and cumulative density function (CDF), being $z_t = (\rho_t - \mu_t)/\sigma_t$. In this case, $(\mu,\sigma^2)$ are the prediction parameters computed with (\ref{eq:predgp}) and $\rho_t=\max(y_1,\ldots,y_t)$ is the incumbent optimum at that iteration.

\section{ROBUST OPTIMIZATION\label{sec:robust}}


Robust optimization is the field of optimization that deals with uncertainty in the parameters of the problem itself or its solution. Specifically, local robustness guarantees that the optimality of the solution is valid even after small perturbations of the solution. Usually, these perturbations are defined in terms of a valid set or region. However, in robotics, perturbations such as location error are represented with probabilistic distributions. Thus, instead of selecting the point that optimizes a single outcome, we select the point that optimizes an \emph{integrated outcome}:
\begin{equation}
\label{eq:integrated}
    g(\x) = \int_\cX f(\x) p(\x) d\x = \ex_{p(\x)} [f(x)]
\end{equation}
where $f(\x)$ is the expected utility from equation \eqref{eq:reward} and $p(\x)$ corresponds to the probability associated with the local perturbations. It can be interpreted objectively as noise on the input variables $\x$ or, subjectively, as a probabilistic representation of the local stability or safety region. That is, a region that guarantees good results even if the query is repeated several times. Instead of $f(\cdot)$, the \emph{integrated outcome} $g(\cdot)$ becomes the function that will be optimized. For the remainder of the paper, we assume that the perturbations (or safety region) are normally distributed, that is, $p(\x) = \mathcal{N}(\mathbf{0}, \Sigma_\x)$. 

Classical robust optimization, considering the worst case scenario, has been previously studied in the context of Bayesian optimization \cite{bogunovic2018adversarially}. Input noise has been addressed to find narrow optima despite query perturbations \cite{oliveira2019bayesian}. 

In this paper, we present an integrated response method based on the unscented transformation, which serves as a cheap and scalable numerical integration method. A preliminary version of this method was published in Nogueira et al. \cite{nogueira2016unscented}. In this case, we use a flexible variant of the unscented transformation, called the scaled unscented transformation \cite{Julier02ACC}, to allow more control on the stability region and avoid numerical issues. Furthermore, previous preliminary work \cite{nogueira2016unscented} interpreted $p(\x)$ as a subjective safety region of stability without actual input noise during training. In this case, we consider the more challenging scenario of both stability and input noise, where the objective is not only to find a broad maximum, but queries are also perturbed $\x \pm \Delta\x$ during training.


\subsection{Scaled unscented Bayesian optimization} 
The unscented transformation is a method to propagate probability distributions through nonlinear transformations with a trade off between computational cost and accuracy. 
The unscented transformation uses a set of deterministically selected samples from the original distribution (called \emph{sigma points}) and transforms them through the nonlinear function $f(\cdot)$. Then, the transformed distribution is computed based on the weighted combination of the transformed sigma points:
\begin{equation}
\mathcal{X}_{0:d} = \left\{\bar{\x}, \; \bar{\x} \pm \left(\sqrt{(d+\gamma)\Sigma_\x}\right)_i \right\} \qquad \forall \; i = 1 \ldots d
\label{eq:sigmapoints}  
\end{equation}
where $(\sqrt{\cdot})_i$ is the i-th row or column of the corresponding matrix square root, $\gamma = \alpha^2 (2d + \kappa) - d$ and $\Sigma_\x$ is the covariance matrix of the perturbation probability distribution. The weight for the initial point is $w^0 = \frac{\gamma}{d+\gamma}$ and $w^{(i)} = \frac{1}{2(d+\gamma)}$ for the rest. The parameters should follow $\kappa \geq 0$ and $0\leq \alpha \leq 1$, while the standard unscented transformation is a special case for $\alpha = 1$. As pointed out by van der Merwe \cite{vdMerwe04Thesis}, we recommend a $\kappa=0$ and $\alpha$ close to $1$. For the matrix square root function, we use the Cholesky decomposition for its numerical stability and robustness \cite{vdMerwe04Thesis}.

Using the unscented transform, we can approximate $\ex_{p(x)}[f(x)] \approx \sum_{i=0}^{2d} f(\mathcal{X}_i) w^i$. The unscented transformation is used twice in our algorithm. First, we need to drive the queries towards points where the improvement is very likely within the safety or perturbed region. Thus, we apply the unscented transformation to the acquisition function, resulting in the \emph{unscented expected improvement}, that is,  $UEI(x) = \sum_{i=0}^{2d} EI(\mathcal{X}_i) w^i$. Intuitively, the UEI will be large when the expected improvement is also large within a region (for example, if the optimum is flat) and UEI will be small when the expected improvement is high only in a small region, like a narrow optimum. However, when exploring, the algorithm might query and select narrow optima by chance. Thus, the incumbent $\rho_t$ cannot be chosen solely based on observed values because we might end up selecting an unstable solution. Therefore, we further compute the \emph{unscented optimal incumbent} to select the most stable optimum $\rho_t = \max_{\x \in \X} \ex_{p(x)}[f(x)] \approx \max_{\x \in \X} \sum_{i=0}^{2d} f(\mathcal{X}_i) w^i$. Note that, in the context of Bayesian optimization, we want to reduce the number of evaluations of $f(\x)$. Therefore, the sigma points are evaluated in the surrogate model, using the GP mean function as an approximation of the target function $\mu(\x) \approx f(\x)$. For that, we approximate the integrated outcome $\widehat{g}(\x) = \sum_{i=0}^{2d} \mu(\mathcal{X}_i) w^i$, where $\mu(\x)$ is obtained from equation \eqref{eq:predgp}. We define the unscented optimal incumbent as $\widetilde{\rho}_t = \max_{\x \in \X} \widehat{g}(\x)$, that can be used in equation \eqref{eq:ei}. Most importantly, we select the final optimal incumbent $\widetilde{\rho}_T \approx \max_{\x \in \X} \ex_{p(\x)} [f(x)]$ as the output of the optimization process.


\subsection{The Unscented transform as integration} 
The unscented transform can be interpreted as a probabilistic integration method. The unscented transformation with $\alpha = 1$ and $\kappa = 0$ is equivalent to the three point Gauss-Hermite quadrature rule \cite{vdMerwe04Thesis}. While the Gauss-Hermite method computes the integral \emph{exactly} under the Gaussian and polynomial assumptions, it has a cost of $\mathcal{O}(n^d)$ where $n$ is the polynomial order of the function in the region. Meanwhile the unscented transform, has a quadratic cost $\mathcal{O}(d^2)$ for computing the integrated response \cite{vdMerwe04Thesis}. The low cost of the unscented transformation is also an advantage compared to other more advanced integration methods such as Monte Carlo or Bayesian quadrature, which have higher computational cost. Note that, during optimization the integrated outcome $g(\x)$ is always approximated with respect to the Gaussian process mean function $\mu(\x)$ to avoid increasing the number of queries of $f(\x)$. Therefore, the integral would be as accurate as the Gaussian process with respect to the target function. We found that, in practice, it is more efficient to employ the computational resources to improve the surrogate model (for example, using MCMC on the kernel hyperparameters), than to provide a better integrated outcome. 

\section{STATISTICAL ROBUSTNESS}



In this section, we focus on mitigating the effect of surrogate modeling errors. In Bayesian optimization, the exploration and exploitation is guided by the surrogate model. By selecting a specific surrogate model (specific GP kernels, Bayesian neural network architectures...) we introduce some assumptions that the target function might not satisfy. Having input noise during the optimization is another source of misleading observations, as the observed query will deviate from the intended query. In practice, a biased or overconfident model produces limited exploration or erroneous exploitation, resulting in even more biased data. This might result in lack of convergence. We propose using an alternative acquisition method, called \textit{Boltzmann selection}, to enhance exploration and provide statistical robustness.

\subsection{Boltzmann selection\label{sec:offpol}}


As a sequential decision making process, we can interpret the Bayesian optimization framework as a \textit{partially observable Markov decision process} (POMDP) \cite{ToussaintBSG,ijcai2019-0327}. In this interpretation, the state is the target function, the action is the next query point, the belief is the surrogate model and the action-value (Q-function) is the acquisition function for each possible query.  
Note that this POMDP model would represent the actual learning/optimization process during policy search. Therefore, equation \eqref{eq:acquisition} can be interpreted as a \emph{meta-policy}, since it is used on a higher abstraction level to learn the actual robot policy $\pi_\x$. We can see that the Bayesian optimization meta-policies found in the literature are greedy, that is, they select the single action or next query that maximizes the acquisition function or \emph{Q-function}.

Our approach consist on replacing the greedy policy of equation \eqref{eq:acquisition} with a stochastic policy such as the Boltzmann policy (also known as Gibbs or softmax policy):
\begin{equation}
  \label{eq:sp}
  p(\x_{t+1}\;|\;y_{1:t}, \x_{1:t}) = \frac{e^{\beta_t \alpha\left(x_{t+1}, p(f\;|\;y_{1:t},\x_{1:t})\right)}} {\int_{x\in\cX} e^{\beta_t \alpha\left(x, p(f\;|\;y_{1:t},\x_{1:t})\right)} dx}
\end{equation}
which defines a probability distribution for the next query or action \cite{ijcai2019-0327}. Thus, the actual next query is selected by sampling that distribution $\x_{t+1} \sim   p(\x_{t+1}\;|\;y_{1:t}, \x_{1:t})$. This policy allows exploration even if the model is completely biased or overconfident, increasing the performance in the presence of modeling errors. This approach can be applied to any acquisition function or surrogate model that can be found in the literature. Since it relies on sampling from a Boltzmann policy, we refer to this query selection approach as the \emph{Boltzmann selection}.

Theoretical analysis shows that the standard \emph{greedy expected improvement} policy may not converge for unknown GP hyperparameters. Instead, in order to guarantee near-optimal convergence rates, it is necessary a \textit{greedy in the limit with infinite exploration} (GLIE) policy \cite{Bull2011,ijcai2019-0327}. A simple and well-known GLIE policy is the $\epsilon$-greedy strategy. This was the policy used to obtain the convergence rates \cite{Bull2011}, but it is highly inefficient and it is never used in practice. Selecting an appropriate $\beta_t$ sequence, the policy from equation \eqref{eq:sp} is GLIE by construction \cite{ijcai2019-0327}. Thus, the Boltzmann policy used in this work is the first policy based on the \emph{expected improvement} that has both good performance in practice and guaranteed theoretical convergence.

\begin{figure*}
  \smallskip \smallskip
  \centering
  \includegraphics[width=0.32\linewidth]{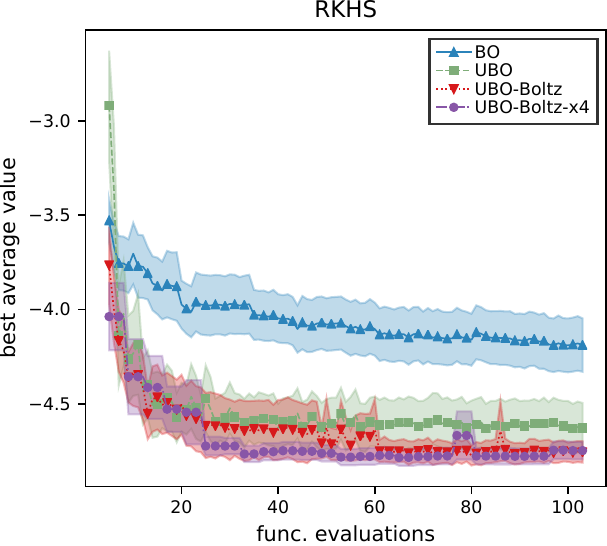}
  \includegraphics[width=0.32\linewidth]{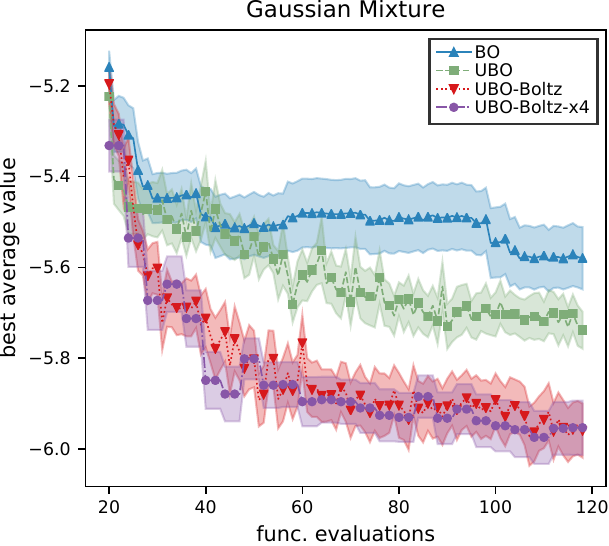}
  \includegraphics[width=0.32\linewidth]{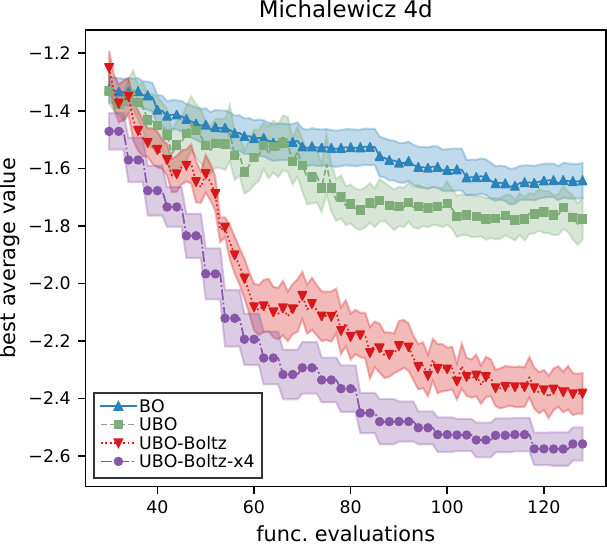}
  \caption{Benchmark functions optimization results. In general, UBO is able to find a more stable solution than the vanilla BO, resulting in a better average value. However, using Boltzmann selection results in an improved stability. Parallelized runs had a much lower walltime without a penalty in performance.}
  \label{fig:benchmark}
\end{figure*}






\subsection{Distributed Bayesian optimization\label{sec:distributed}} 


A secondary advantage of using the Boltzmann selection is that they also trivially enable distributed optimization, where different policy parameters can be evaluated in parallel in a fully distributed fashion. This could be applied in multi-robot scenarios or for simulation-based reinforcement learning. Many parallel methods for Bayesian optimization have been proposed in the past few years with heuristics to enforce diverse queries. Some authors include artificially augmented data by hallucinated observations \cite{ginsbourger2010kriging,Snoek2012}, combine optimization with some degree of active learning in order maximize the knowledge about the target function \cite{desautels2014parallelizing,contal2013parallel,shah2015parallel} or enforce spatial coverage \cite{gonzalez2016batch}. All these methods have in common the need for a central node that shares and synchronizes the data and the queries to enforce diversity in the parallel runs.

Sampling using this Boltzmann selection already ensures diverse queries with random numbers \cite{ijcai2019-0327}. Thus, a centralized node is not required and all computation can be done in each node in a fully distributed manner. In terms of communication, the nodes only need to broadcast their latest evaluated query and observation value $\{\x_{t}, y_t\}$, requiring minimal communication bandwidth. Communication can be asynchronous and it is robust to delays or failures in the network, as the order of the queries and observations is irrelevant.


\subsection{Baseline surrogate model}
Surrogate model selection introduces some assumptions that might not be satisfied in the problem at hand. This might result in a biased model as discussed before. This effect can be mitigated by choosing an expressive and flexible model, compatible with Bayesian optimization. In Gaussian processes, model expressiveness is related to the choice of kernel, which also encodes the prior information about the function space. The most frequent kernels in Bayesian optimization are stationary kernels $k(x,x') = k(|x-x'|)$ like the Mat\'ern kernel. However, robotic and reinforcement learning settings typically showcase nonstationarity \cite{MartinezCantin17icra}, therefore a nonstationary kernel is needed. Consider again the example from Figure \ref{fig:rover_example}. Given that the optimum can be either trajectory depending on the trajectory uncertainty level, our optimization algorithm must be able to model both if needed. Thus, it must be able to model the shape or both narrow and broad optima in different regions. This is known to be problematic in Bayesian optimization. For that reason, we have incorporated the Spartan kernel \cite{martinez2018funneled} as presented in Section \ref{sec:bayesianoptimization}. Furthermore, we estimate the kernel hyperparameters using MCMC. Contrary to maximum likelihood or maximum a posteriori estimates which tend to underestimate the model uncertainty, MCMC allows a more flexible model. Note that both unscented Bayesian optimization and the Boltzmann selection are agnostic of the kernel or estimation algorithm choice. By selecting a flexible surrogate as a baseline, we highlight the importance of Boltzmann selection even with limited bias.

\section{RESULTS}

In this section we describe the experiments used to compare the performance of different Bayesian optimization methods in the presence of input noise. We compare a vanilla implementation of Bayesian optimization (BO), the unscented Bayesian optimization with a greedy policy (UBO) and the unscented Bayesian optimization with the Boltzmann selection (UBO-Boltz). We also compare with a parallel version of the Boltzmann selection with 4 nodes (UBO-Boltz-x4) to study the performance impact of adding parallelization. Note that in the results we show the number of function evaluations, not iterations. For example, at the 20 evaluation, the UBO-Boltz-x4 method had run only for 5 iterations, therefore requiring less wall time and using only the previous information from 16 trials instead of 19.

\begin{figure*}
  \smallskip \smallskip
  \centering
  \includegraphics[width=0.24\linewidth]{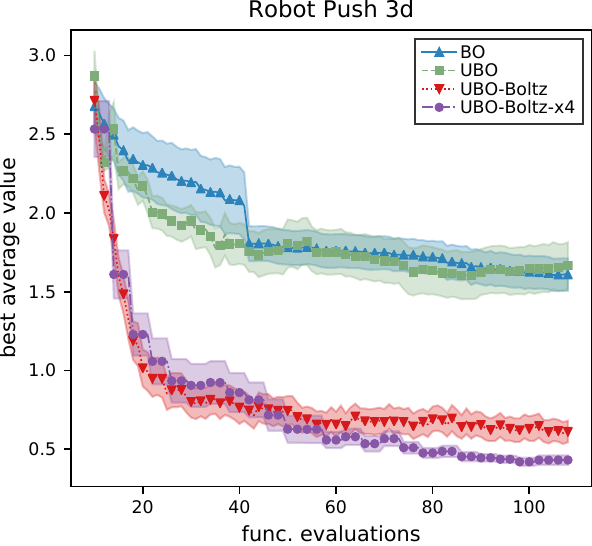}
  \includegraphics[width=0.24\linewidth]{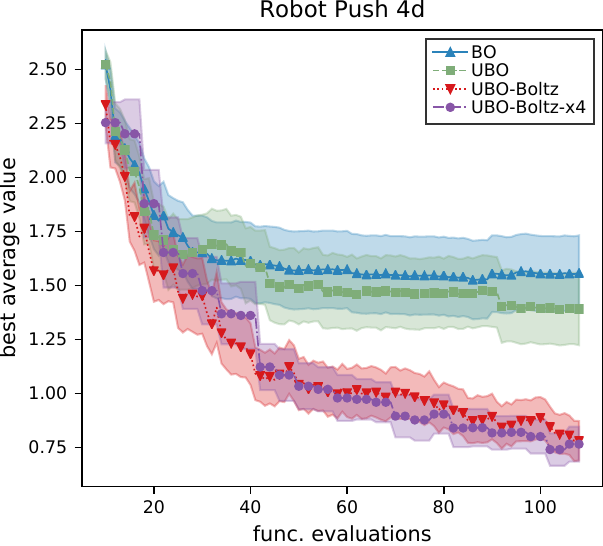}
    \includegraphics[width=0.24\linewidth]{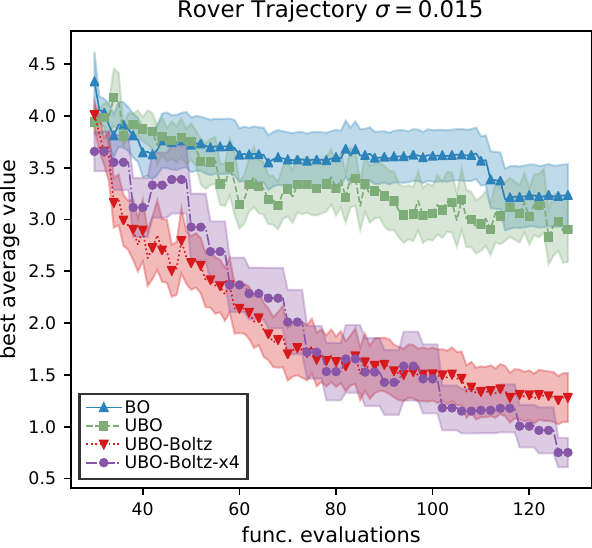}
  \includegraphics[width=0.24\linewidth]{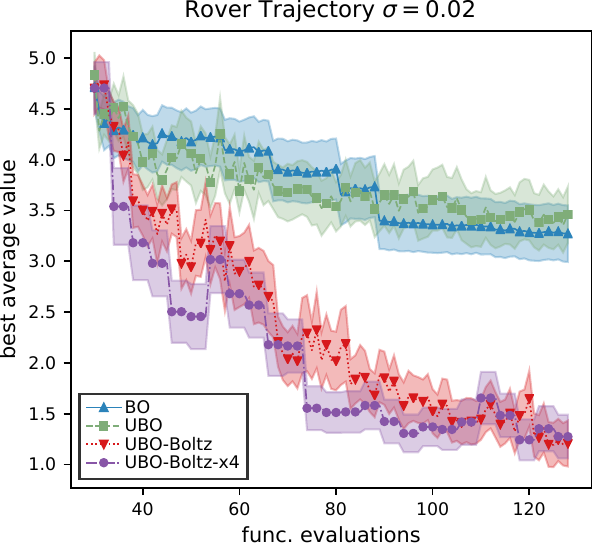}
  \caption{Robot pushing problem and rover path planning optimization results. For the more complex problems, the UBO is not able to find a stable solution, unlike the Boltzmann selection.} 
  \label{fig:robotpush}
\end{figure*}

All methods share the same configuration as a baseline: expected improvement (EI) as the acquisition function and a Gaussian process as the surrogate model with the Spartan kernel and MCMC for the hyperparameters. Given that EI is known to be unstable during the first iterations due to lack of information \cite{Jones:1998,Bull2011}, the optimization is initialized with $p$ evaluations from a low discrepancy Sobol sequence.

The performance of each method was evaluated in the following way: For every function evaluation $\x_{t}$, each method computes their best solution (the optimal incumbent $\rho_t$ or the unscented optimal incumbent $\widetilde{\rho}_{t}$) using the observations $(\x_{1:t},y_{1:t})$ and according to their model at that point of the optimization. Then, we evaluate the integrated outcome at the best solution $g(\widetilde{\x}_{t})$ by approximating \eqref{eq:integrated} using 1000 Monte Carlo samples from $p(\x)$ over the actual function $f(\cdot)$. For the plots, we repeat each optimization 20 times and display the mean value with 95\% confidence interval. Common random numbers were used for all the methods. We assume isotropic input noise, reported as $\sigma$ such that $\Sigma_x = I\sigma^2$ from \eqref{eq:sigmapoints}. The input space is normalized on all the problems between 0 and 1, so the reported input noise $\sigma$ is already normalized.


\subsection{Benchmark Optimization Functions}

We have evaluated the methods on synthetic benchmark functions for optimization. We have used the functions previously used in the BO literature: the RKHS function \cite{freitas2015} and a Mixture of 2D Gaussian distributions (GM) \cite{nogueira2016unscented}. These functions have unstable global optima for certain levels of input noise. This means that in order to locate the safe optima, we need to model and take into account the input noise. We have also used a 4D Michalewicz function\footnote{https://www.sfu.ca/~ssurjano/optimization.html}, a popular test problem for global optimization because of its sharp edges and the large number of local optima. All benchmark functions use input noise $\sigma=0.02$ and 100 evaluations. The number of initial samples is set based on the dimensions of each problem to 5, 20 and 30 samples for RKHS, GM and Michalewicz.

Figure \ref{fig:benchmark} shows the results on the benchmark functions. Although UBO finds better stable optima than BO, using the Boltzmann selection further improves the performance. It also shows that adding parallelization barely impacts the optimization results. This means that we can achieve better performance and wall-time using the parallel approach.

\newcommand\roverwidth{0.95} 
\begin{figure}
    \centering
    
        {\scriptsize \rotatebox[origin=c]{90}{{BO}}} 
        \raisebox{-0.5\height}{\includegraphics[width=\roverwidth\linewidth]{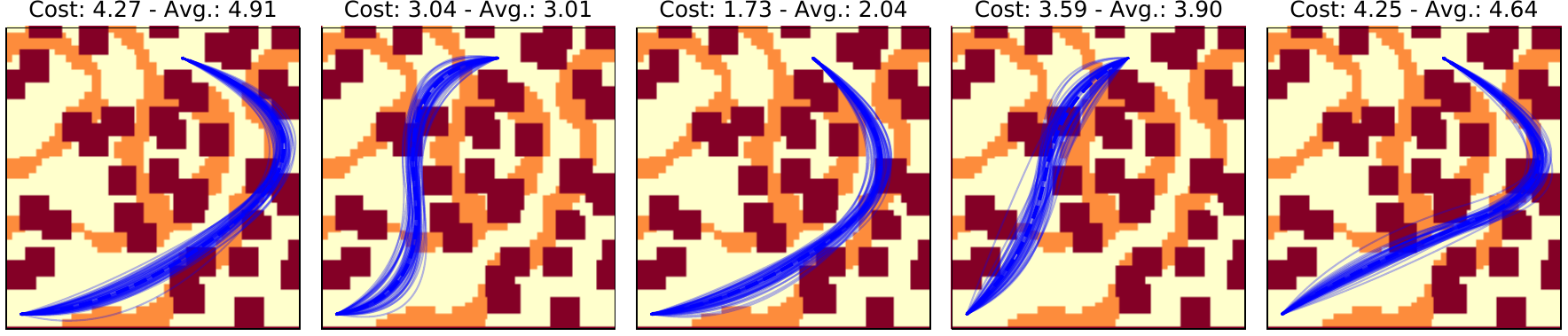}} \\ 
        {\scriptsize \rotatebox[origin=c]{90}{{UBO}}} 
        \raisebox{-0.5\height}{\includegraphics[width=\roverwidth\linewidth]{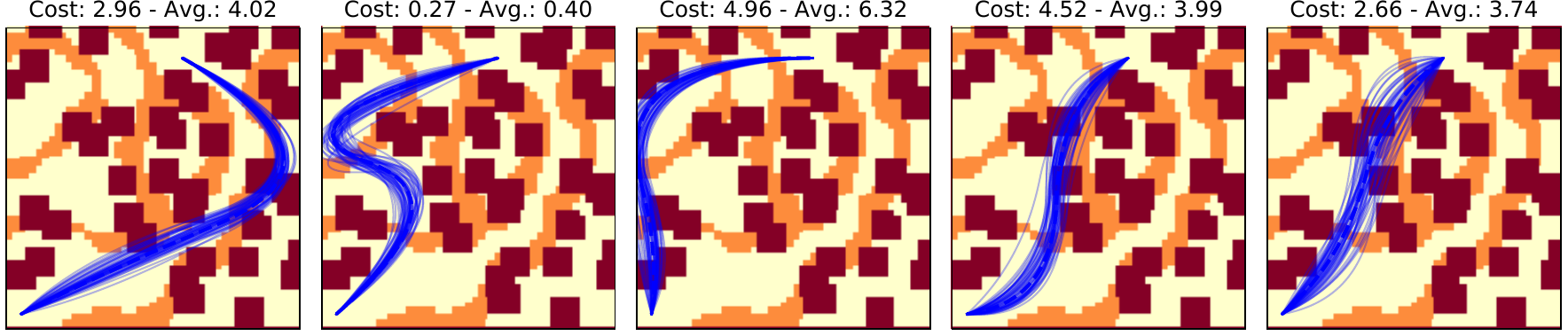}} \\
        {\scriptsize \rotatebox[origin=c]{90}{{UBO-Boltz}}} 
        \raisebox{-0.5\height}{\includegraphics[width=\roverwidth\linewidth]{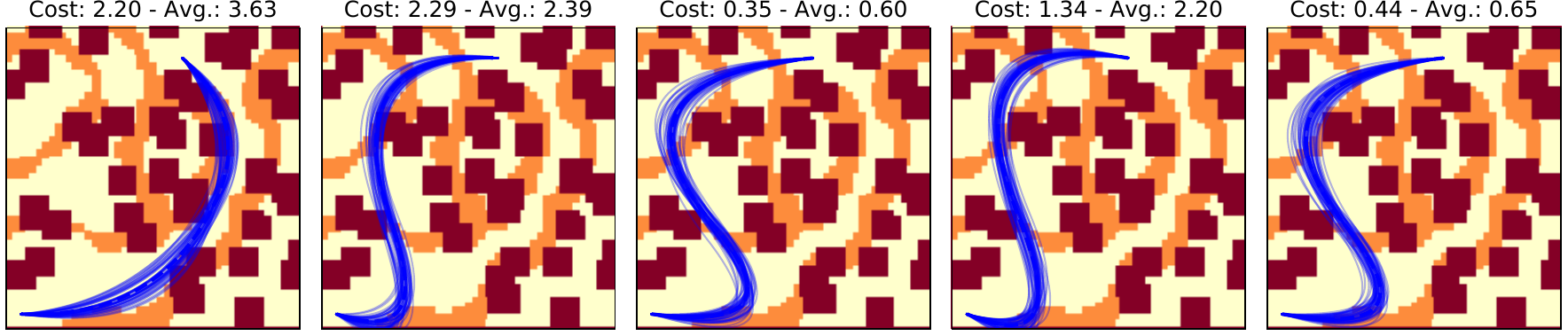}} \\
        {\scriptsize \rotatebox[origin=c]{90}{{UBO-Boltz-x4}}} 
        \raisebox{-0.5\height}{\includegraphics[width=\roverwidth\linewidth]{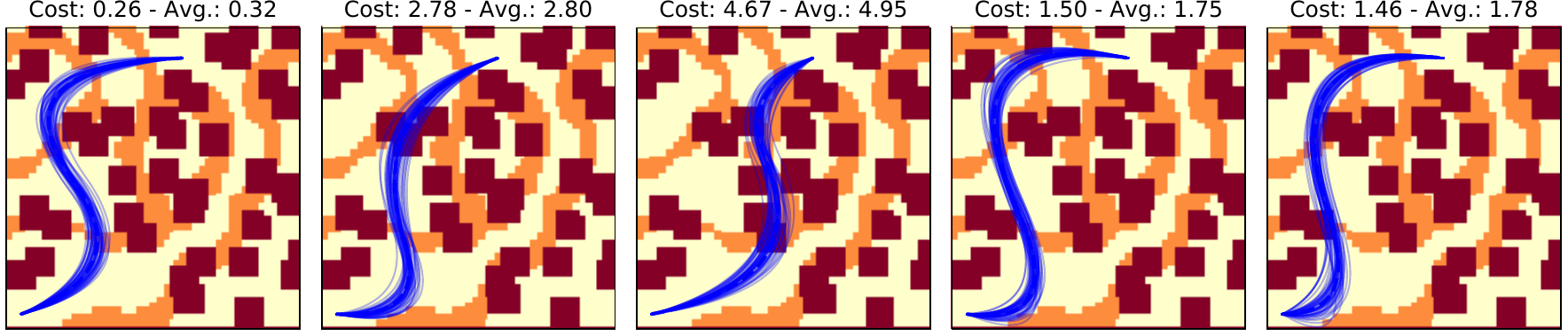}} \\
    
    \caption{Examples of optimized trajectories found by different methods (rows) and trials (columns), showing the possible deviations from the trajectories by simulating input noise $\sigma=0.02$. We display the cost of the desired trajectory (assuming no input noise) and the average cost from possible deviations over each result. Note that BO (first row) does not find the safe path in any of the trials. The best trial is found by UBO (2nd row and column: noiseless cost 0.27, noisy cost 0.40), however most of the time lacks exploration to find a good one. The Boltzmann selection (serial in row 3 and parallel in row 4) improves exploration allowing better results overall.}
    \label{fig:roverresults}
\end{figure}

    
    

\subsection{Robot Pushing}


Next, we have used the \emph{active learning for robot pushing} setup and code from Wang et al. \cite{pmlr-v70-wang17e}. The task is to push an object towards a designated goal location. In the 3D problem, the policy parameters are the robot location $(r_x, r_y) \in [-5,5]$ and pushing duration $t_r \in [1,30]$, while pushing direction $r_\theta$ towards the object is assumed to be known. In the 4D problem, $r_\theta \in [0, 2\pi)$ is included as the fourth parameter. The problem is simulated in a 2D environment with a circular robot and a square object. The robot follows a constant linear motion from $(r_x, r_y)$ in $r_\theta$ direction during $t_r$ timesteps and the simulation is run until both object and robot reach zero velocity. The policy cost is the distance between the object and the designated goal. In both functions, we use 10 initial queries and a budget of 100 function evaluations during optimization. The 3D version uses $\sigma=0.02$ while the 4D version uses $\sigma=0.01$. We reduced the input noise in the 4D function because the robot angle parameter $r_\theta$ is very sensitive to input noise, as a small change in direction of the robot might result in completely missing the goal.

Figure \ref{fig:robotpush}, shows the Robot Pushing results. Contrary to benchmark results, UBO no longer guarantees stable solutions. Instead, it shows that Boltzmann selection is required to achieve good results. Parallel performance remains similar.

\subsection{Robot Path Planning}

In this section we cover the problem of safe path planning. The objective is to find a stable and efficient trajectory of a rover through a rugged terrain with multiple obstacles. It is based on \textit{optimizing rover trajectories} from Wang et al. \cite{wangAISTATS2018}. In this case, there are 4 policy parameters within the $[0,1]$ interval, a pair of 2D reference points for a trajectory defined by a quadratic B-Spline with smoothness condition set to 0 (to enforce it to go through the reference points). The robot has to perform path planning while avoiding obstacles, which might be dangerous for the rover to collide with, and steep slopes, which might be dangerous as the rover can tip over. The cost for traversing obstacles or getting out of bounds is 20.0, for slopes is 1.0 and a constant cost of 0.05 for penalizing trajectory length. The overall policy cost is computed by integrating along the trajectory. In the figures, obstacles are red rectangles and slopes are orange regions. We are interested in finding stable trajectories that avoid the danger that might arise from trajectory deviations. This is a common problem in robot navigation as localization errors might result in the robot not following the desired trajectory accurately \cite{MartinezCantin07RSS,MartinezCantin09AR}.

We study this problem using 2 different input noises: $\sigma = 0.015$ and $\sigma=0.02$. We use 30 initial samples and 100 function evaluations during optimization. Figure \ref{fig:robotpush} shows the resulting optimization performance of each method. Figure \ref{fig:roverresults} shows some trajectories obtained using different methods. We can see how both BO and UBO are prone to get stuck in suboptimal trajectories as solutions while Boltzmann selection methods return a solution closer to the safe trajectory.

\section{CONCLUSIONS}
In this paper, we propose the first active policy search algorithm that is robust both in a statistical and optimization point of view. For achieving both types of robustness, we use multiple techniques, such as, the unscented transformation to deal with input noise and local perturbations, a Boltzmann selection to enhance the exploration and convergence guarantees of the acquisition function. Therefore, this paper presents several contributions that provide synergistic results. First, we have presented a new formulation and interpretation of the unscented Bayesian optimization algorithm and shown its robustness both for safety/stability conditions and against small perturbations in the queries. Second, for statistical robustness, we have used a Boltzmann selection of the acquisition function for actively preventing surrogate bias and guaranteeing near-optimal convergence. This further highlights previous results that indicates that the ubiquitous greedy strategy in the Bayesian optimization literature can be suboptimal in many applications. In the experiments, we have used an an adaptive Gaussian process with the Spartan kernel for modeling the nonstationarity of reward/cost functions in robotic applications, to guarantee a strong and flexible baseline. The method has been evaluated on several benchmark functions and robotic applications which showcase the influence of input noise, such as safe robot navigation. We also take advantage of the embarrassingly parallel nature of the Boltzmann selection that could be used in multi-robot setups or simulation environments.




%
%





\bibliographystyle{IEEEtran}
\bibliography{IEEEabrv,references,ijcai19}


\end{document}